\pdfoutput=1

\documentclass[11pt]{article}

\usepackage[]{EACL2023} 

\usepackage{times}
\usepackage{latexsym}

\usepackage[T1]{fontenc}

\usepackage[utf8]{inputenc}

\usepackage{microtype}

\usepackage{inconsolata}

\usepackage{wrapfig}
\usepackage{hyperref}
\usepackage{graphicx}
\usepackage{multirow}
\usepackage{hyperref}
\usepackage{bm}
\usepackage{seqsplit}
\usepackage{xcolor}
\usepackage{stmaryrd}
\usepackage{booktabs}
\usepackage{fixltx2e}
\usepackage{xparse}
\usepackage{amsmath}
\usepackage{amssymb}
\usepackage{amsfonts}
\usepackage{soul}
\usepackage{arydshln}

\DeclareGraphicsExtensions{.png,.pdf}
\label{packs}

\NewDocumentCommand{\blue}{ }{\textcolor{blue}}

\NewDocumentCommand{\heng}{ mO{} }{\textcolor{red}{\textsuperscript{\textit{Heng}}\textsf{\textbf{\small[#1]}}}}

\NewDocumentCommand{\chenkai}{ mO{} }{\textcolor{blue}{\textsuperscript{\textit{Chenkai}}\textsf{\textbf{\small[#1]}}}}

\NewDocumentCommand{\cheng}{ mO{} }{\textcolor{purple}{\textsuperscript{\textit{Cheng}}\textsf{\textbf{\small[#1]}}}}

\NewDocumentCommand{\tie}{ mO{} }{\textcolor{teal}{\textsuperscript{\textit{Tie}}\textsf{\textbf{\small[#1]}}}}

\newcommand\blfootnote[1]{%
  \begingroup
  \renewcommand\thefootnote{}\footnote{#1}%
  \addtocounter{footnote}{-1}%
  \endgroup
}

\title{Incorporating Task-Specific Concept Knowledge into Script Learning}

\author{First Author \\
  Affiliation / Address line 1 \\
  Affiliation / Address line 2 \\
  Affiliation / Address line 3 \\
  \texttt{chenkai5@illinois} \\\And
  Second Author \\
  Affiliation / Address line 1 \\
  Affiliation / Address line 2 \\
  Affiliation / Address line 3 \\
  \texttt{email@domain} \\}

\author{Chenkai Sun, Tie Xu, {\bf ChengXiang Zhai}, {\bf Heng Ji} \\
         University of Illinois at Urbana-Champaign, IL, USA\\
         Alibaba, Hangzhou, China\\
\texttt{\{chenkai5, czhai, hengji\}@illinois.edu} \\
\texttt{xutie.xt@alibaba-inc.com} \\
}

\begin{document}
\maketitle

\begin{abstract}

In this paper, we present \textsc{Tetris}, a new task of Goal-Oriented Script Completion. Unlike previous work, it considers a more realistic and general setting, where the input includes not only the goal but also additional user context, including preferences and history. To address this problem, we propose a novel approach, which uses two techniques to improve performance: (1) concept prompting, and (2) script-oriented contrastive learning that addresses step repetition and hallucination problems. On our WikiHow-based dataset, we find that both methods improve performance. \blfootnote{\url{https://github.com/chenkaisun/Tetris}}

\end{abstract}

\section{Introduction}
\label{sec:intro}

A Goal-Oriented Script refers to a sequence of events that describe some stereotypical activities for achieving a specified goal \cite{ai_handbook, gosc}. It is important to study how to 
automatically construct instructional scripts because it enables 
many high-impact applications such as robot action planning~\cite{robot_action_planning,robot_motion_planning}, causal reasoning~\cite{causal_survey}, and task-oriented dialogue generation \cite{dialogue_survey}. While previous work has shown that neural models are capable of constructing the entire script given a goal~\cite{gosc,proScript}, their proposed tasks are highly restrictive and based on overly simplified assumptions about the application because it ignored both the usage context  %
(e.g., what is a preferred way of solving the problem, and what steps have been executed already) and the variable personal preferences that a user may have (e.g., a goal might need be achieved in different ways).  Take ``Make Eggless Cupcakes'' as an example, the user might want to make it with different equipment (e.g., by using a rice cooker instead of an oven) or styles (e.g., to have some banana flavor), or the user has already completed some steps but would like to seek information on what to proceed next; a machine learning model trained on goal-to-script tasks as done in the existing work cannot adaptively provide solutions under such situations. 


\begin{figure}
	\vskip 0.2in
	\centering
	\includegraphics[width=1\linewidth]{"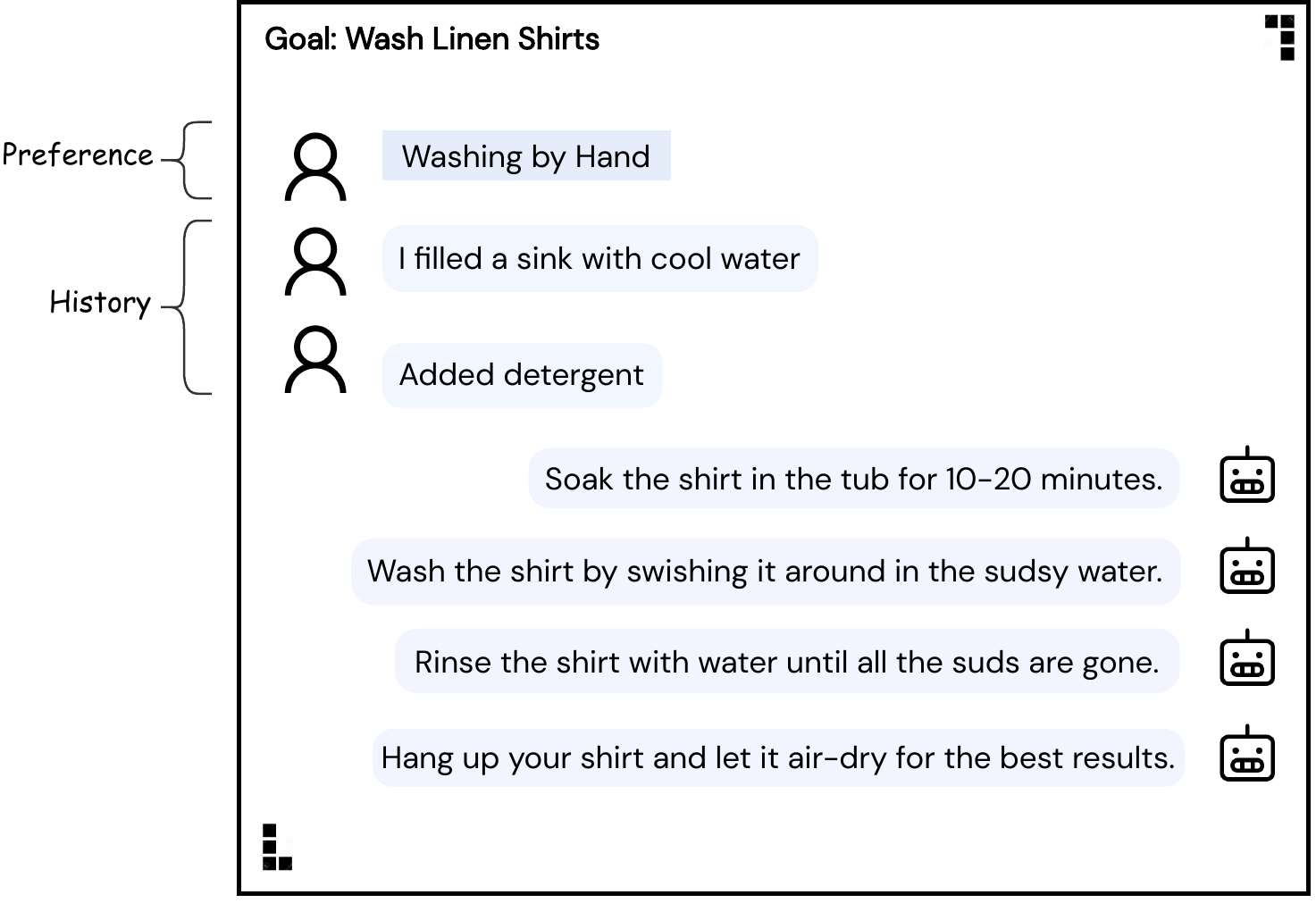"}
	\caption{An example illustrating the task. The input consists of a specified goal (e.g., wash line shirts), an optional preference (e.g., washing by hand), and a history of steps. The model is asked to generate the remaining steps to achieve the goal.}
	\label{fig:example}
	\vskip -0.2in
\end{figure}
We address the limitations of the previous task formulation and propose \textsc{Tetris}, a more general task of goal-oriented script learning that allows flexible usage scenarios, and therefore it accommodates more realistic needs in downstream applications. More specifically, as shown in Figure~\ref{fig:example}, the task considers both user preference and history as input in addition to the goal.  From a probabilistic perspective, instead of modeling the problem with $p$(Script|$\mathcal{G}$), we model it as 

	\vskip -0.2in
\begin{gather*}
		p(\text{Completion of Script}|\mathcal{H, P, G})  \\  = \prod_i p(\text{Script}_i|\text{Script}_{<i},\mathcal{P, G})
\end{gather*} 

where $i$ indicates each step index in the script, $\mathcal{P}$ denotes user preference, $\mathcal{H} $ indicates History, and $\mathcal{G}$ abbreviates Goal.

\begin{figure*}
	\centering
	\includegraphics[width=0.9\linewidth]{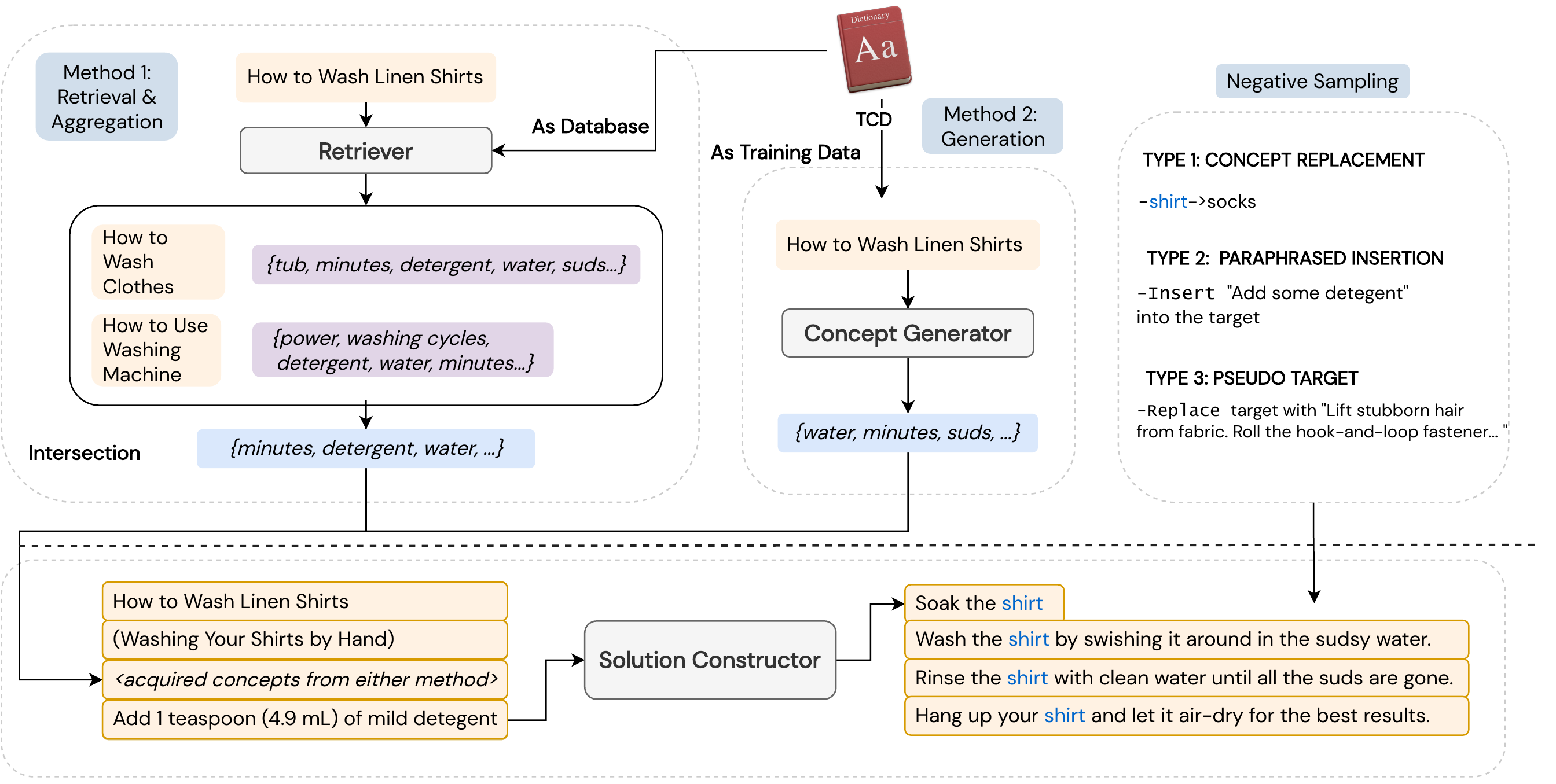}
	\caption{Our goal-oriented script completion framework. The top part shows the methods for automatically acquiring relevant concepts from the Task Concept Dictionary and negative sampling strategies. The bottom part shows the solution constructor, which uses the acquired concepts as an informative prompt to complete the script. }
	\label{fig:method}
\end{figure*}

To solve the new problem, we can use a commonly used baseline method, i.e., using a seq2seq model \cite{bart, t5}, but a direct application of such a model would not be optimal because the model's understanding of the context of the goal is inevitably shallow and thus unlikely be able to fully exploit the extra context information we have in the problem setup. To address this limitation, we propose to enrich the representation with additional task-specific concepts so as to enable the model to understand the context more deeply and leverage relevant knowledge about those concepts encoded in the training data the model has been exposed to. To obtain task-specific knowledge needed for solving script learning, we introduce \textsc{Task Concept Dictionary} (TCD), a novel Key-Value Knowledge Base (KB) that consists of task phrases as the key and the associated concepts involved in the solution process for each task as the value. The concepts contain not only items needed for the task but also concerned attributes (e.g.,  ``thickness'') and intermediate products during the process (e.g., ``dough'' appears during Making Rice Noodles). In our work, as a proof of concept, we automatically construct the knowledge base from WikiHow\footnote{\url{www.wikihow.com}}, one of the most used and actively updated How-to websites.
Once we construct a TCD, it can be potentially used for solving the problem of script completion in many ways. In this paper, we focus on exploring how to leverage it to acquire task-specific relevant concepts for enriching the task representation used in a baseline model for script completion. 
Specifically, inspired by how humans use mind-map to construct solutions, we introduce a novel "concept prompting" framework that first acquires relevant concepts from TCD and then connects them with the input as a prompt for the language generator to complete the script (Figure~\ref{fig:method}). We further introduce two complementary ways of concept acquisition. The first method retrieves concepts of close neighbors from TCD. The second method involves a 
generative model pretrained on TCD that is capable of generating associated concepts for a given goal, which can generalize beyond similar cases. Discovering that the model often repeats historical steps and hallucinates actions and entities, we also developed a script-oriented contrastive learning approach to address these issues by constructing corresponding negative samples.

%
We perform experiments on a dataset constructed from WikiHow and find that our method gives a consistent improvement on both automatic and human evaluations in comparison to the baseline, demonstrating the benefit of TCD for script completion. Moreover, we find that concept acquisition quality has a large impact on this task.

To summarize, we make the following contributions:
\begin{itemize}
    \item We introduce \textsc{Tetris}, the task of Goal-Oriented Script Completion that asks a model to complete the task based on a goal, and optionally preference and history. The task poses interesting new challenges and enables new applications. 
    
    \item We propose \textsc{Task Concept Dictionary}, a novel task-specific knowledge base that encodes knowledge about the associations between goals and concepts and a method composed of (1) extraction and integration of the relevant concepts from TCD, and (2) script learning-oriented contrastive learning objective to enhance the correctness of model-produced scripts.


    
    
    \item We experiment with the proposed methods 
    on the WikiHow dataset and show that our methods outperform the state-of-the-art baseline models consistently, demonstrating the need for task-specific knowledge and the benefit of TCD for improving the performance of script completion. 
\end{itemize}

\section{Task Formulation}


We define a Goal-Oriented Script to include three types of elements, \textcolor{purple}{\textbf{Goal}}, \textcolor{teal}{\textbf{Preference}}, and \textcolor{darkblue}{\textbf{Step}}. A \textcolor{purple}{\textbf{Goal}} represents the task desired to be completed. An optional \textcolor{teal}{\textbf{Preference}} defines in what ways the goal should be completed. A \textcolor{darkblue}{\textbf{Step}} is one procedural event toward solving the task, conditioned on a preference. In this work, given a goal $\mathcal{G}$, a preference $\mathcal{P}$, and a history of steps $\mathcal{H}$ already done, we aim to generate the remaining steps that accomplish the goal in natural language. An example is shown in Figure~\ref{fig:example}. For convenience, we use $\mathcal{G}_p$ to denote the goal conditioned on $\mathcal{P}$, and $\mathcal{I}$ to denote the entire input (consisting of $\mathcal{G}$, $\mathcal{P}$, and $\mathcal{H}$).


\section{Task Concept Dictionary}
As in virtually all NLP applications, a main technical challenge in solving the problem of script completion is how to acquire the relevant knowledge and leverage the knowledge to generate the remaining steps in a script. In our case, a critical question is: What kind of knowledge representation is likely helpful for solving the script completion problem? In our work, we make a hypothesis that creating task-concept association would help the model generalize better on script completion and propose Task Concept Dictionary (TCD). In TCD, each task key $\mathcal{G}$ from the collection of the keys $\mathcal{D}^G$ (e.g., ``How to Modify the Navigation System of an Acura'') is mapped to a set of relevant concepts $\mathcal{D}^\mathcal{G}_C$ during its solution process. Each $c\in\mathcal{D}^\mathcal{G}_C$ can be associated with $\mathcal{D}^\mathcal{G}$ in four types of roles of a concept in that task: as preparation material needed for achieving the task (like ingredients for recipe), as attributes/aspects that need to be considered, as intermediate entities generated during the process, and others. In this work, we made a primal construction of TCD (TCD.v0) as a proof of concept, that is, the association types are not considered between $\mathcal{G}$ and $\mathcal{D}^\mathcal{G}_C$. 

The design of TCD is motivated by the observation that humans can often form solutions through brainstorming and mind-mapping with  knowledge about
the tasks and relevant concepts involved in the solution process. Analogously, TCD is meant to encode such task-specific knowledge about associations between goals and concepts so as to assist downstream script learning applications. As will be shown in our experiment results, TCD is indeed beneficial for improving performance for script completion by supplying useful relevant concepts to a baseline model. 



In general, TCD can be constructed by using any instructional content on the Web. In our experiments, to construct TCD.v0, we first collect all articles from WikiHow. 
We chose to use WikiHow as the seed corpus for its richness in user preference data.
Each article consists of a goal, a list of preferences, and solution steps under each preference. Let $\mathcal{G}^{p_j}_i$ be the goal from an article $i$ conditioned on the $j$-th preference, we then proceed to augment the set $\mathcal{D}^G$  with a key $\mathcal{G}^{p_j}_i$, and augment $\mathcal{D}^{\mathcal{G}^{p_j}_i}_C$ with noun phrases (by using Spacy\footnote{\url{https://spacy.io/}} tagger for extraction) in the solution steps under preference $j$. We create an association between goals and concepts if they co-occur in the same article. We show the statistics of TCD in Appendix~\ref{sec:dataset}.

\section{Method}
\label{sec:method}

In this section, we discuss our design of the method for \textsc{Tetris}. The framework contains two parts, concept acquisition from TCD and solution construction. The first part uses TCD as an aid to acquire concepts relevant to the current input. Specifically, we introduce two different methods (Figure~\ref{fig:method}), one based on retrieval of closely related tasks and aggregation on the associated concepts, while the other involves training a concept generator on TCD. As shown in Section~\ref{sec:experiment}, both of them have a positive impact on performance, yet they also introduce different benefits. The concepts fetched from TCD using either method form a concept prompt, which is combined with the input $\mathcal{I}$ (consisting of goal, preference, and history) as an 
augmented input $\mathcal{I}^*$. $\mathcal{I}^*$ is then fed into an encoder-decoder language model $\mathcal{M}_{\text{EncDec}}$ to complete the script. Below we describe the two methods for the generation of relevant concepts using TCD. 
\subsection{Concept Retrieval \& Aggregation}
\label{sec:method1}
The intuition of the retrieval method comes from observing how human refers to their past knowledge or the web for similar instructional sources (even though they might not address the need exactly) to achieve the current goal. For instance, if someone has experience in baking cheesecake and chocolate cake, it would be a breeze for them to make a strawberry cake by adapting the knowledge from their past experience.  Building on the intuition, we propose to use retrieval as an interpretable way to retrieve similar tasks from the set of keys $\mathcal{D}^G$ in TCD and use their associated concepts to aid the downstream script completion task. 



\noindent\textbf{Retrieval.} First we encode $\mathcal{G}_p$ (i.e., the preference-conditioned goal in $\mathcal{I}$) into a dense vector $\mathbf{e}_\text{g}$ with an encoder model. We similarly encode $\mathcal{D}^G$ into $\{\mathbf{e}_j\}_{j=1}^{|\mathcal{D}^G|}$. A cosine similarity score $s_{gj}$ is computed between $\mathbf{e}_\text{g}$ and each $\mathbf{e}_\text{j}$. The top-$K$ related tasks $\mathcal{N}^K$ are then retrieved based on the scores. We further obtain a set of concepts  $\mathcal{C}_i$ for $i$-th task $\mathcal{N}^K$. In our experiment, we use SBERT~\cite{sbert} (pretrained for semantic search) as the encoder. We use FAISS~\cite{faiss}, an efficient similarity search library, to perform top-k search.  


\noindent\textbf{Aggregation.} The retrieved neighborhood of concepts, however, may sometimes introduce contextual noise (e.g., the grape and chocolate-related concepts are not useful in  strawberry cake baking). To tackle this issue, we additionally perform operations on the retrieved concepts using the set intersection. The concept set for an input $\mathcal{I}$ is computed as $\mathcal{C}^s=\bigcap_{i=1}^{i=K} \mathcal{C}_i$. We finally map the set $\mathcal{C}^s$ into a list $\mathcal{C}$, neglecting ordering information.

\begin{table}[t]
\begin{small}
\centering
\begin{tabular}{llll}
\toprule

 & F\&E & F\&B & C\&V \\ \hline\noalign{\vskip 0.7ex}
 
\# Train Samples & 10600 & 2824 & 2214 \\
\# Dev Samples & 3449 & 873 & 741 \\
\# Test Samples & 3567 & 930 & 714 \\
\# Articles & 2201 & 802 & 481 \\
Avg \# Tokens/Step & 8.66 & 7.25 & 9.31 \\
Avg \# Steps/Article & 10.42 & 8.66 & 10.41 \\\bottomrule
\end{tabular}
\caption{Statistics summarizing the WikiHow-based dataset for \textsc{Tetris}, where F\&E indicates Food and Entertainment, F\&B indicates Finance and Business, and C\&V indicates Cars and Other Vehicles}
\label{tab:data-stats}

	\vskip -0.1in
\end{small}
\end{table}

{\renewcommand{\arraystretch}{1} 

\begin{table*}[t]
	\centering
	\begin{small}

    \begin{tabular}{cllllllllll}
    \toprule
    \textbf{Method}  & \textbf{BERTScore} & \textbf{BARTScore} & \textbf{BLEU 1} & \textbf{BLEU 2} & \textbf{BLEU 3} & \textbf{BLEU 4} &\textbf{METEOR} &  \textbf{ROUGE 2} \\
    \midrule
    BART & 86.76 & -4.65 & 8.45 & 4.40 & 2.21 & 1.13 & 14.67 & 4.6 \\
    GPT2 &86.75 &-4.59 &17.89 &8.29 &3.44 &1.52 &15.61 &3.1
 \\\hdashline\noalign{\vskip 0.5ex}

CRA-3 & 87.18 & -4.41 & 22.59 & 10.95 & 4.86 & 2.10 & 18.07 & 4.9 \\
CRA-1 & 86.74 & -4.50 & 21.87 & 10.82 & 4.85 & 2.21 & 17.55 & 5.3 \\
CG & 86.86 & -4.45 & 24.34 & 11.84 & 5.26 & 2.43 & 17.88 & 5.1 \\
CG+SOCL & 86.77 & -4.47 & \textbf{26.18} & \textbf{13.15} & \textbf{6.18} & \textbf{3.07} & 18.11 & \textbf{5.7} \\
CRA-2 & \textbf{87.21} & \textbf{-4.37} & 23.95 & 11.36 & 5.06 & 2.35 & \textbf{18.65} & 4.9 \\
CRA-2+SOCL & 87.19 & -4.38 & 24.25 & 11.58 & 5.25 & 2.42 & 18.36 & 5.0 \\
    \bottomrule 
    \end{tabular} \newline\newline 
    \begin{tabular}{cllllllllll}
    \toprule
    \textbf{Method}  & \textbf{BERTScore} & \textbf{BARTScore} & \textbf{BLEU 1} & \textbf{BLEU 2} & \textbf{BLEU 3} & \textbf{BLEU 4} &\textbf{METEOR} &  \textbf{ROUGE 2} \\
    \midrule

    BART & 87.87 & -4.37 & 16.61 & 8.31 & 4.67 & 2.92 & 18.92 & 5.6 \\
    
    GPT2 &  88.06 &-4.30&17.13 &7.27 &3.58& 2.10& 18.61& 3.6\\\hdashline\noalign{\vskip 0.5ex}
CRA-3 & 88.37 & -4.28 & 14.73 & 6.95 & 3.77 & 2.36 & 19.17 & 4.6 \\
CRA-1 & 87.66 & -4.37 & 17.68 & 8.57 & 4.63 & 2.87 & 18.95 & 5.4 \\
CG & 88.35 & \textbf{-4.17} & 22.87 & 10.97 & 6.09 & 3.94 & 20.88 & 5.4 \\
CG+SOCL & \textbf{88.44} & -4.18 & \textbf{23.42} & \textbf{11.12} & \textbf{6.19} & \textbf{4.02} & \textbf{21.20} & 5.7 \\
CRA-2 & 88.28 & -4.26 & 18.78 & 9.00 & 4.94 & 3.18 & 19.73 & 5.2 \\
CRA-2+SOCL & 88.42 & -4.20 & 20.48 & 10.04 & 5.80 & 3.96 & 20.89 & \textbf{6.3} \\

 \bottomrule 
    \end{tabular}\newline
    
    \begin{tabular}{cllllllllll}
    \toprule
    \textbf{Method}  & \textbf{BERTScore} & \textbf{BARTScore} & \textbf{BLEU 1} & \textbf{BLEU 2} & \textbf{BLEU 3} & \textbf{BLEU 4} &\textbf{METEOR} &  \textbf{ROUGE 2} \\
    \midrule
BART & 87.95 & -3.88 & 28.39 & 14.88 & 7.76 & 4.30 & 21.99 & 7.3 \\
GPT2 & 87.88& -3.94& 22.73 &11.06 &5.43& 2.91 &19.95& 5.7\\

\hdashline \noalign{\vskip 0.5ex}
CRA-3 & 88.05 & -3.89 & 26.43 & 14.18 & 7.62 & 4.32 & 21.39 & 7.4 \\
CRA-1 & 87.94 & \textbf{-3.85}  & 29.09 & 15.07 & 7.76 & 4.18 & 22.05 & 7.4 \\
CG & 87.84 & -3.90 & \textbf{29.69} & \textbf{15.70} & \textbf{8.27} & \textbf{4.60} & 21.96 & 7.4 \\
CG+SOCL & 87.92 & -3.87 & 27.97 & 14.66 & 7.81 & 4.42 & 21.93 & \textbf{7.6} \\
CRA-2 & 88.06 & \textbf{-3.85}  & 28.48 & 14.71 & 7.64 & 4.17 & 21.89 & 7.3 \\
CRA-2+SOCL & \textbf{88.08} & \textbf{-3.85} & 29.40 & 15.54 & 8.21 & 4.58 & \textbf{22.41} & 7.5 \\

 \bottomrule 
    \end{tabular}
    
    \caption{Automatic metrics for evaluating language generation performance on the test set of Vehicles (top), Finance (middle), and Food (bottom). Our main models \textbf{CRA-2} and \textbf{CG} outperform the baseline consistently, demonstrating the effectiveness of our mechanism. We also include the performance of \textbf{CRA-1} and \textbf{CRA-3} to show that the performance varies as the number of neighbors changes. Moreover, we show that our contrastive learning objective (\textbf{SOCL}) almost always helps the model to have better performance.}
	\label{table:test}
    \end{small}
\end{table*}

}

\subsection{Concept Generation}
\label{sec:method2}

While the retrieval method allows an explicit concept acquisition process, it is limited by the width of TCD. For example, if one would like to complete a task from a domain that is not covered well by TCD, the retrieved neighbors can introduce more noise than help. To address this limitation of the retrieval method and enhance generalization, we propose to use language modeling as an alternative way of acquiring concepts.
Specifically, inspired by the work~\cite{comet}, in which a concept is generated given an edge type,
we propose to directly generate a set of concepts relevant to a given task and preference. We directly train the model on TCD, where the model is asked to generate the set $\mathcal{D}^t_C$ given each key $t$. In the inference stage, by feeding $\mathcal{G}_p$ into the trained concept generator, we can then obtain a list of concepts $\mathcal{C}$.



\subsection{Solution Constructor}

In this step we aim to encode both of the information from $\mathcal{C}$ (using either of the above methods) and $\mathcal{I}$ to generate the remaining steps that accomplish the goal specified in $\mathcal{I}$. The TCD has enabled us to enrich the representation by augmenting $\mathcal{I}$ with $\mathcal{C}$ and such an augmented representation can then be fed into any baseline script completion method to enable the baseline method to have (indirect) access to the knowledge encoded in TCD, thus improving accuracy of script completion.  





\noindent \textbf{Input Formation}. An encoder-decoder model $\mathcal{M}_g$ is used for generation. In our experiment, we choose BART~\cite{bart} as the base model for its impressive performance on other NLP tasks \cite{kgbart,bart_use}. Clearly, the framework accommodates any other models as well. Given the list of concepts $\mathcal{C}$ generated from either Section \ref{sec:method1} or \ref{sec:method2}, we form the input $\mathcal{I}_f$  to the encoder as 
``\texttt{<s>}Goal (Preference) \#\#\# $\mathcal{C}$ \#\#\# \texttt{</s>}Step$_{1}$\texttt{</s>}Step$_{2}$...Step$_{|\mathcal{H}|}$\texttt{</s>}'', where \texttt{<s>} is a special token to represent the start of the sentence and \texttt{</s>} is for separation. 


\noindent \textbf{Model Generation}. The input is fed through the standard tokenization, embedding mapping, and transformer to produce encoder hidden states. In the decoder end, the decoder hidden states (from the previous token) additionally attend to encoder hidden states to produce the next token. The model computes generation probability by taking the dot product between the decoder output and the tokens embeddings from the vocabulary.

Lastly, we use negative log-likelihood loss during training for each sample

\vskip -0.2in
 
\begin{equation}
\begin{split}
    \mathcal{L}_{G}=-\sum_{i=1}^{|S|} \log P\bigg(s_i|s_{<i}, T\bigg)
\end{split}
\end{equation}
 
 where $S$ denotes the tokens for the remaining steps.

\subsection{Script-Oriented Contrastive Learning}








To further improve the accuracy of the generated scripts, we design a contrastive learning framework that addresses deficiencies discovered in the model outputs from \ref{sec:method1} and \ref{sec:method2}.



\noindent \textbf{Negative Sampling} In contrastive learning\cite{cl2}, hard negative samples are constructed to guide the model to better distinguish between the incorrect samples and the desired outcome. To construct such negatives, we gain insights from the model output, from which we found that the model elicits two typical types of erroneous behavior: (1) repetition of steps from the history and (2) hallucination of non-relevant actions/concepts for a given task (e.g., sago appeared in the process of making kimchi). Based on the observation, we propose the following strategies: (1) \textsc{Concept Replacement}, where we randomly replace the concepts in the positive sample with concepts from other tasks in TCD under the same category (e.g., Food category in WikiHow), (2) \textsc{Paraphrased Insertion}, where we paraphrase history steps and insert them into target steps, and (3) \textsc{Pseudo Targets}, where we construct pseudo target by sampling steps from the same category and glue them into a sequence. 

\noindent \textbf{Contrastive Loss} 
In computing the contrastive loss, we generate negative targets from the strategy above for each positive script target (composition discussed in the Appendix). To allow the model to distinguish between correct and incorrect script completions, we use the model's hidden states to compute a score to compute the correctness score for each sample, which is then used to compute the triplet loss~\cite{triplet_loss}. More specifically, for each training sample

	\vskip -0.1in
$$
\begin{aligned}
&\mathcal{L}_{\mathrm{CL}}=\sum_k \max \left(0, \phi+c_{k}^{-}-c^{+}\right), \\
&c^{+}=\sigma\left(\operatorname{AvgPool}\left(\boldsymbol{W}_{c} \boldsymbol{H}^{+}+\boldsymbol{b}_{c}\right)\right) \\
&c_{k}^{-}=\sigma\left(\operatorname{AvgPool}\left(\boldsymbol{W}_{c} \boldsymbol{H}_{ k}^{-}+\boldsymbol{b}_{c}\right)\right)
\end{aligned}
$$
	\vskip -0.1in

Where $\boldsymbol{H}^{+}$ and $\boldsymbol{H}_{ k}^{-}$ indicate the decoder hidden states for the positive and $k$-th negative sample, $\sigma$ is the sigmoid function, AvgPool is the average pooling function, $\boldsymbol{W}_{c}$ and $\boldsymbol{b}_{c}$ are learnable parameters.


\subsection{Training Objective}

We jointly optimize the model on cross-entropy loss from generation and triplet loss from contrastive learning $$\mathcal{L}=\mathcal{L}_{G}+\beta\mathcal{L}_{CL}$$ where $\beta$ is hyper-parameter.

\section{Experiments}
\label{sec:experiment}

\subsection{Dataset}

We collect data from Food \& Entertaining (abbreviated as Food), Finance \& Business (abbreviated as Finance), and Cars \& other Vehicles (abbreviated as Vehicles) categories of WikiHow, which have varying data scales and therefore allow comparison of models' generalization ability. Each article from WikiHow contains a goal, (optionally) several preferences for completing the goal, and steps under each direction. The details of the dataset are shown in table~\ref{tab:data-stats}. More processing details are in Appendix~\ref{sec:dataset}. 







\subsection{Implementation and Training Detail}

Our  model  is  implemented  using Pytorch~\cite{pytorch}  and  Huggingface Transformers~\cite{huggingface} with BART-base as the base generator. The reproducibility and hyperparameter details can be found in Appendix~\ref{sec:impl}.


\subsection{Compared Methods}
We compare our framework and methods with the state-of-the-art text generation baselines \textbf{BART}~\cite{bart} and \textbf{GPT2}~\cite{gpt2}, which don't use any of our proposed methods. The solution constructor in our framework also uses BART. We use \textbf{CRA}-$k$ to denote top-$k$ neighbors in \ref{sec:method1} and set $k$ to be 1,2, or 3 in our experiment. We use \textbf{CG} to denote the  concept generator in \ref{sec:method2}. Since TCD.v0 is also based on WikiHow as source data, we make some modifications to our methods in the experiment so that we can test the scenarios where the new unseen task cannot be exactly found in the knowledge base. For concept generation, we exclude the evaluation set related articles in the dataset from its training data. For \textbf{CRA}-$k$, when we retrieve relevant tasks from TCD for a given goal, we remove its own key from TCD. \textbf{SOCL} indicates that script-oriented contrastive learning is used during the training.

\begin{table*}[t]
	\begin{small}
\centering
\begin{tabular}{llllll}
\toprule
\textbf{Name} & \textbf{Correctness ↑} & \textbf{Fluency ↑} & \textbf{Average Rank ↓} & \textbf{Best ↑} & \textbf{Worst ↓} \\  \midrule

BART & 1.72 & 2.57 & 3.57 & 4.13 & 72.73 \\
CG & 2.32 & 3.06 & 2.65 & 9.09 & 13.22 \\
CRA-2 & \textbf{2.52} & \textbf{3.15} & \textbf{2.34} & \textbf{14.88} & \textbf{9.92}\\ \midrule
Gold-Concepts & \textbf{3.40} & \textbf{3.55} & \textbf{1.44} & \textbf{71.90} & \textbf{4.13}  \\  \bottomrule\\
\end{tabular}
\caption{Human evaluation results for the WikiHow dataset on all three categories combined. Scripts completion outputs (from automatic analysis) from each model are presented to human judges, who are asked to rate on Correctness (i.e., at what level does the generated solve the problem) and Fluency of the output. We also report the average rank and percentage of the time that each model output is chosen as best or worst, after asking each judge to rank outputs from different methods for each input}
\label{tab:human-eval}
    \end{small}
    \vskip-15pt
\end{table*}


\subsection{Automatic Evaluation}
\label{sec:autoeval}
\noindent \textbf{Evaluation Metrics} We use both deep learning and n-gram-based evaluation strategies.
The metrics include BERTScore~\cite{bertscore}, BARTScore~\cite{bartscore}, BLEU~\cite{bleu}, METEOR~\cite{meteor}, and ROUGE~\cite{rouge}. Note that BARTScore computes the log-likelihood of producing the reference text given the generated text using a BART model pretrained on ParaBank2\footnote{\url{https://github.com/neulab/BARTScore}}. BERTScore computes an embedding matching-based score.

\noindent \textbf{Results} The test set results are shown in Table~\ref{table:test}. We can see that the performance of our method variants (both CRA and CG) is consistently better than the baselines, empirically demonstrating the effectiveness of both concept acquisition methods and that the task-concept representation of the task-specific knowledge is helpful for downstream script learning. Moreover, while CG outperforms CRA-2 on all metrics in Finance, it is less clearly observed in the others, showing that the level of effectiveness also depends on the domain. The fact that CRA-2 is better than CRA-1 most of the time shows that the set operation in Section~\ref{sec:method1} does improve the retrieval quality for certain domains. Meanwhile, CRA-3 consistently performs worse than CRA-2, showing that setting $k=3$ may have removed too much useful information. With contrastive learning, the model almost always gains enhancement in performance, showing the effectiveness of our sampling strategies; an exception is shown in the Food category for CG, where it shows positive impact only on half of the metrics, and this might be caused by that the CG itself is effective enough and contrastive learning introduces more noise than help. We also show in Appendix that adding SOCL alone is already effective. Lastly, we observe that the Vehicle and Finance categories contain much less training data than Food, yet the performance improvement is more notable, demonstrating the potential of the methods in low-resource settings. 

\subsection{Human Evaluation}


\noindent \textbf{Evaluation Metrics \& Process} As mentioned in ~\cite{ccb_tutorial}, automatic metrics do not correlate well with human judgments on script learning tasks due to the diversity of potential solutions. As a complement to automatic metrics, we also recruited five volunteers (who are not authors) and conducted the human evaluation. The volunteers were Master/Ph.D. students with enough background knowledge to rate output. We gave each candidate a quiz composed of 20 random samples from the dataset as filtering to see if the annotator can give scores close to the authors'. We made sure that each annotator understands the assigned data samples during evaluation. In our evaluation process, we compared the output generated by BART, CRA-2, and CG with SOCL, and Gold-Concepts (a variant where concepts come groundtruth in TCD). We presented the input and the corresponding generated outputs from each method to human judges and asked them to rate the Correctness and Fluency of the output, both on a scale from 0 to 4. Correctness (or usefulness) was defined as the level of confidence that the generated steps successfully complete the specified goal and preference. We additionally asked the judges to rank the outputs according to their overall preference and their rating on Correctness \& Fluency; after this procedure, we reported the average rank and the percentage of the time that each model output is chosen as the best or worst. 



\noindent \textbf{Results}. 121 samples are randomly selected from automatic analysis for human evaluation, where 40 come from the Vehicle category, 40 come from Finance, and 41 from Food. We present the human evaluation results in Table~\ref{tab:human-eval}. The results align with automatic metrics roughly in general. The Gold-Concept variant is usually the best since it has access to ground-truth noun phrases from the reference. CRA-2 often generates more helpful solutions for users. The outputs from BART are 
worse than others most of the time, confirming the belief that using acquired task knowledge as prompts can help the model to generate better solutions on average. The Fluency level for all methods is rated to be of acceptable quality most of the time, likely due to the strong generalization capability of the pretrained language model.

\begin{table}[t]
	\begin{small}
\centering

\begin{tabular}{ll}
\toprule
\textit{Goal} & \textbf{How to Store Peaches} \\ \midrule
\textit{Preference} & Keeping Peaches in the Fridge \\ \midrule
\textit{History} & \begin{tabular}[c]{@{}l@{}}Rinse the peaches to clean off any \\ dirt or debris. Dry the peaches with \\ clean paper towels or a clean hand towel.\end{tabular} \\ \midrule
BART & \textless{}eos\textgreater{} \\ \midrule
CG & \begin{tabular}[c]{@{}l@{}}{[}\textit{peach, paper bag, store peach,} \\ \textit{container, term storage}{]}\\ \textbf{\textcolor{blue}{Store peaches in the fridge}} for up to \\ 3 months.\end{tabular} \\ \midrule
CRA-2 & \begin{tabular}[c]{@{}l@{}}{[}\textit{peach, freezer, container, half}{]}\\ \textbf{\blue{Store the peaches in an airtight container}} \\ in the freezer for up to 3 months.\end{tabular} \\ \midrule
Reference & \begin{tabular}[c]{@{}l@{}}Place whole, uncut peaches in the fridge \\ on their own or in a plastic bag. ...\\ \textbf{\textcolor{red}{Store sliced peaches in an airtight}} \\ \textbf{\textcolor{red}{container}} for 1-2 days.\end{tabular} \\ \bottomrule
\end{tabular}

\caption{Example input and output from the Food category. The noun phrases contained in square brackets in CG and CRA-2 indicate the acquired concepts from TCD. <eos> indicates an empty completion (i.e., the history is believed to contain the complete script). With fridge storage-related concepts, our models correctly produce the storing step. }

\label{tab:gen-example-food2}
    \end{small}
    \vskip-15pt

\end{table}


\begin{figure}[h]
	\centering
	\includegraphics[width=0.9\linewidth]{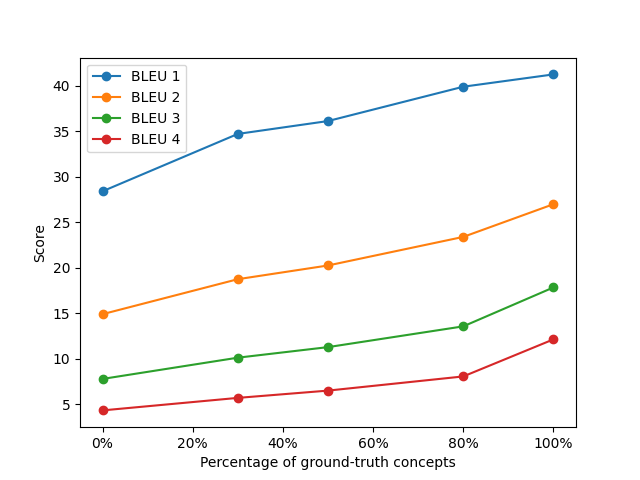}
	\caption{The plot shows that the automatic performance increases as the amount of ground-truth concepts increases, showing that deriving the right task-specific concepts can make the model generate better scripts.}
	\label{fig:goldperc}
\end{figure}
\subsection{Impact of Retrieval Quality} 


One question to be asked is how impactful it is to derive a concept prompt close to the groundtruth (i.e., concepts from TCD by using the current goal as the search key). To answer the question, we randomly draw concepts from the groundtruth at different thresholds and use each of them as a prompt to train a baseline model. From our result shown in Figure~\ref{fig:goldperc}, we can see that the performance monotonically increases as the number of concepts covering the ground-truth concepts increases, showing the consistent benefit of using the concepts in the right context; this also demonstrates the promising direction of our methods and the significance of developing a concept deriver better at understanding task context.

\subsection{Qualitative Analysis}
\label{sec:quala}
We present an example of generated outputs by different methods in Table~\ref{tab:gen-example-food2}. Additional examples are available in Appendix~\ref{sec:model-output}. We compare BART, CG, and CRA-2. From the generated output, we can see that BART ends the script immediately, missing the step of storing, likely due to the lack of access to knowledge about storage.  On the other hand, with the acquisition from TCD, our methods can reach more contextual knowledge for the current task. Specifically, both methods are able to access storage-related concepts such as term storage, container, and freezer. As a result, the outputs match with the reference more closely. While the alignment with the reference isn't perfect (e.g., peaches are additionally placed in a bag in the reference), to most human judges, the script completion clearly achieves the goal already; this further demonstrates the difficulty of evaluating script completion.

\section{Related Work}
\label{sec:related}

\noindent {\bf Script Learning:}
Scripts~\cite{schank2013scripts, ai_handbook, visual_gosc,gosc2,gosc3} refers to the knowledge of stereotypical event sequences which human is constantly experiencing  and  repeating  every day. One branch of works in script learning focuses on distilling \textit{narrative} scripts from news or stories \cite{mcnc, mcnc2, mcnc3}, where the scripts are not goal-oriented. One of the most recent tasks in \textit{narrative} script modeling is Multiple-Choice Narrative Cloze Test~\cite{mcnc7}, where an event is removed from the chain and the model is asked to predict which event from the choices fills the blank. The other line of work, which is more closely related to our work, centers around \textit{procedural} scripts, where a sequence of events happened often to achieve a goal. Recent work has introduced the task of constructing the entire script given a goal \cite{gosc} or choosing 1 out of 4 candidates' steps that most likely help achieve a goal~\cite{ccb_reason}. In this work, we propose a more general script learning setup that considers additionally the user preference and history, improving the generalization of models in script learning.

Our method is related to Retrieval-augmented text generation \cite{rag_survey,ra_dialog,ra_dialog2}, a paradigm that aims to combine deep learning models with retrieval methods for text generation and has gained significant attention in recent years. For example in~\cite{prompt_ret}, the author retrieves similar training examples as prompt to the current input. In our work, we are the first to introduce a framework that retrieves concepts from the concept dictionary as prompts for script learning. Apart from acquiring concepts, we also introduce a way to cancel contextual noise from neighbors by set intersections for better prompt quality. Furthermore, our method is related to contrastive learning~\cite{cl1,cl2,cl3}, a line of self-supervised learning methods that improve representation learning by compacting positive samples while contrasting them with negative samples~\cite{cl4}. Previous methods used negative sampling approaches such as replacing the words with synonyms and shuffling target sentences. In our work, we introduced a new script-oriented contrastive learning objective to address script-specific issues and enhance the quality of generated scripts.

\section{Conclusion}
\label{sec:conclusion}
In this work, we propose \textsc{Tetris}, the new task of Goal-Oriented Script Completion, which allows a model to produce the remaining steps given a user-specified goal, preference, and history of steps. We also present \textsc{Task Concept Dictionary} (TCD), a knowledge base representing task and concept association, to enable knowledge-based methods for the task. We introduce different methods to acquire concepts as prompts for the downstream text generator. We also introduce a contrastive learning strategy for script learning. The methods present consistently better performance on both automatic and human evaluation, clearly demonstrating the benefit of TCD for improving the performance of script completion. The qualitative analysis further shows 
that the task-specific knowledge can indeed benefit goal-oriented script learning tasks by feeding relevant knowledge about task completion. Future work could explore how to improve the acquisition quality from TCD and applications of TCD on task-oriented dialog systems.

\section*{Limitations}
\label{sec:limitation}

While TCD paired with concept acquisition methods can aid downstream script learning tasks, it doesn't consider the inclusion of actions of each step event, which can potentially benefit the script learning tasks. A possible direction is to extend the design of TCD and the concept prompt to include the semantics of actions and their orders.  

Meanwhile, the concepts extracted by our method do not overlap with the ground-truth concepts (i.e., the set of concepts that appear in the reference) very well (e.g., <20\% in Jaccard Index). The gap in performance between our methods and the Gold-Concept variant shows that improving the concept derivation quality might be the next step.

Furthermore, because our dataset is constructed from the English version of WikiHow, the benefits of our methods shown in the experiments are only empirically proved to work for English. We plan to further test our methods in multiple languages.


%
%



\section*{Ethics Statement}
\label{sec:ethic}

The study aims to extend deep learning-based models on the ability to generalize scripts under different user contexts. The script learning models introduced in the work can potentially be helpful for task-oriented dialog systems to suggest solutions to users. The dataset on which we base our experiments is constructed automatically from the publicly available website WikiHow. Since the website is primarily crowdsourced, the models trained on the data might incur subjective bias. During the human evaluation phase of the experiment, all involved human judges participated voluntarily and received decent payment.

\section*{Acknowledgements}

We would like to thank the anonymous reviewers for their constructive suggestions on our work and thank the raters for their help on manual evaluation. This research is based upon work supported in part by U.S. DARPA KAIROS Program No. FA8750-19-2-1004. The views and conclusions contained herein are those of the authors and should not be interpreted as necessarily representing the of ficial policies, either expressed or implied, of DARPA, or the U.S. Government. The U.S. Government is authorized to reproduce and distribute reprints for governmental purposes notwithstanding any copyright annotation therein.

\bibliography{anthology,custom}
\bibliographystyle{acl_natbib}

\appendix

\section{Appendix}
\label{sec:appendix}

{\renewcommand{\arraystretch}{1} 

\begin{table*}[t]
	\centering
	\begin{small}

    \begin{tabular}{cllllllllll}
    \toprule
    \textbf{Method}  & \textbf{BERTScore} & \textbf{BARTScore} & \textbf{BLEU 1} & \textbf{BLEU 2} & \textbf{BLEU 3} & \textbf{BLEU 4} &\textbf{METEOR} &  \textbf{ROUGE 2} \\
    \midrule
    BART & \textbf{87.87} & -4.37 & 16.61 & 8.31 & 4.67 & 2.92 & 18.92 & 5.6 \\
BART+SOCL & 87.725 & \textbf{-4.349} & \textbf{19.478} & \textbf{9.441} & \textbf{5.216} & \textbf{3.236} & \textbf{19.198} & \textbf{5.73} \\
    \bottomrule 
    \end{tabular} 
    
    \caption{On the Finance dataset, our result shows that SOCL alone already brings consistent benefits to the performance.  }
	\label{table:test_bart}
    \end{small}
\end{table*}

}



\begin{table}
	\caption{Hyperparameters for non-contrastive learning (CL) models (some are introduced and changed for CL experiment). The ones below the mid-line are generation related. Batch size  is changed for contrastive learning.}
	\centering
	\begin{tabular}{ll}
		
		\toprule
		Name &Value\\
		\midrule
		seed &42  \\
		learning rate & 3e-5  \\
		batch size & 16  \\
		weight decay & 5e-4  \\
		RAdam epsilon & 1e-8  \\
		RAdam betas & (0.9, 0.999)  \\
		scheduler & linear  \\
		warmup ratio (for scheduler) & 0.06  \\
		number of epochs & 25  \\
		metric for early stop & SacreBLEU\footnote{\url{https://github.com/mjpost/sacrebleu}} \\
		patience (for early stop) & 15  \\ \hline
		length penalty & 1.2  \\
		max length & 511  \\
		min length & 2  \\
		beam search size & 5  \\
		\bottomrule
	\end{tabular}
	\label{tab:hparam}
\end{table}

\begin{table*}[t]
\centering
\begin{tabular}{ll}
\hline
\textit{Goal} & \textbf{How to Report Pet Store Neglect} \\ \hline
\textit{Preference} & Reporting to Local Law Enforcement \\ \hline
\textit{History} & \begin{tabular}[c]{@{}l@{}}Document the situation as extensively as possible \\ Check your state's animal welfare law \\ Find out which agency is responsible for enforcing animal laws\end{tabular} \\ \hline
BART & Finished \\ \hline
CG & \begin{tabular}[c]{@{}l@{}}[dog] \\ Report the situation to local law enforcement \\ Follow up with any follow-up questions \\ Finished\end{tabular} \\ \hline
CRA-2 & \begin{tabular}[c]{@{}l@{}}[police, report, animal control, authority]\\Contact local law enforcement agencies if necessary\\ Follow up with any follow-up questions\\ Finished\end{tabular} \\ \hline
Gold-Concept & \begin{tabular}[c]{@{}l@{}}[situation, state, animal welfare law, agency, animal law,\\ documentation, record, report, law enforcement]\\Gather documentation and record of the report to law enforcement \\ Follow up with your report if necessary \\ Finished\end{tabular} \\ \hline
Reference & \begin{tabular}[c]{@{}l@{}}Gather your documentation and other records \\ File your report with law enforcement \\ Follow up on your report \\ Finished\end{tabular} \\ \hline
\end{tabular}
\caption{Example output among different methods (Finance and Business)}
\label{tab:gen-example-1}
\end{table*}

\begin{table*}[]
\centering
\begin{tabular}{ll}
\hline
\textit{Goal} & \textbf{How to Make Macaroni Pasta} \\ \hline
\textit{Preference} & Making Pasta Dough at Home \\ \hline
\textit{History} & \begin{tabular}[c]{@{}l@{}}Mix the dry ingredients\\ Create a hole or well in the dry ingredients \\ Add three large eggs \\ Whisk the eggs with a fork \\ Place the dough on a clean counter\end{tabular} \\ \hline
BART & \begin{tabular}[c]{@{}l@{}}Bake the pasta \\ Serve the pasta while it is still warm \\ Finished\end{tabular} \\ \hline
CG & \begin{tabular}[c]{@{}l@{}}[flour, salt, bowl, butter, milk, egg, flour mixture, dough, surface]\\ Roll the dough out into a rectangle \\ Cut the dough into 1/4 inch (0.64 cm) circles \\ Transfer the dough to a plate \\ Repeat the process with the rest of the dough \\ Cook the pasta \\ Place the pasta on a plate to warm up \\ Serve \\ Finished\end{tabular} \\ \hline
CRA-2 & \begin{tabular}[c]{@{}l@{}}[dough] \\Roll the dough into a ball \\ Repeat the process with the rest of the dough \\ Transfer the dough to a baking sheet \\ Bake the pasta \\ Serve or refrigerate \\ Finished\end{tabular} \\ \hline
Gold-Concept & \begin{tabular}[c]{@{}l@{}}[ingredient, hole, egg, fork, dough, counter, surface, bowl]\\Place the dough on a floured surface \\ Transfer the dough to a serving bowl \\ Serve \\ Finished\end{tabular} \\ \hline
Reference & \begin{tabular}[c]{@{}l@{}}Fold the dough \\ Look for a smooth surface \\ Place the dough in a clean bowl to rest \\ Finished\end{tabular} \\ \hline
\end{tabular}
\caption{Example output among different methods (Food and Entertaining category). }
\label{tab:gen-example-food}
\end{table*}

\begin{table*}[t]
\centering
\begin{tabular}{ll}
\hline
\textit{Goal} & \textbf{How to Track a Flight in Progress} \\ \hline
\textit{Preference} & Via Online Websites \\ \hline
\textit{History} & \begin{tabular}[c]{@{}l@{}}Gather the information about the flight:\\ Find a flight-tracking website to find out the information about your flight \\ Enter your flight information \\ Watch for plugins that you could need to add to your \\ \ \ browsers functionality to run the website \\ Watch a map come up on your screen showing the \\ \ \  route the plane has taken, as well as the current location \\ \ \  of the plane and the expected route ahead\end{tabular} \\ \hline
BART & Finished \\ \hline
CG & [flyer, airline, website, flight, status, status update] Finished \\ \hline
CRA-2 & [airline, flight, flight number] Finished \\ \hline
Gold-Concept & \begin{tabular}[c]{@{}l@{}}[information, flight, tracking website, flight information, \\plugin, browser functionality, website, map, screen, route, \\plane, location, auto, update feature, way, site, ability] \\ Use the auto-update feature on the way out of the site \\ Allow the site to continue running while you're not using it \\ Check to see if the update feature is working properly\end{tabular} \\ \hline
Reference & \begin{tabular}[c]{@{}l@{}}Figure out if the website you chose has an auto-update feature\\ Look for ways to zoom in on the plane, if this site allows that ability \\ Finished\end{tabular} \\ \hline
\end{tabular}
\caption{Example output among different methods (Cars \& Vehicles category)}
\label{tab:gen-example-cv}
\end{table*}



\subsection{Dataset \& TCD}
\label{sec:dataset}

The dataset statistics are shown in Table~\ref{tab:data-stats}. The raw corpus comes from the 07/20/21 snapshot of WikiHow. We filter out unordered scripts by using both the classification results from \cite{gosc} and the WikiHow section type. We perform a 6:2:2 split on the articles to create a train, development, and test set. We create a history $\mathcal{H}$ in data samples for \textsc{Tetris} by randomly splitting a sequence of steps under each preference into two halves. For TCD, we have 206621 keys in total and 10.37 concepts per key on average.



\subsection{Implementation Details}
\label{sec:impl}

We implement the models using the 4.8.2 version of Huggingface Transformer library\footnote{\url{https://github.com/huggingface/transformers}}\cite{huggingface}. We use the Oct 1, 2021 commit version of the BART-base model (139M parameters) from Huggingface\footnote{\url{https://huggingface.co/facebook/bart-base/commit/ea0107eec489da9597e9eefd095eb691fcc7b4f9}}. The contrastive learning variants has 140M parameters. For SBERT in Section~\ref{sec:method1}, we use the all-mpnet-base-v2 checkpoint from sentence transformer library~\footnote{\url{https://www.sbert.net/}}. We use Huggingface datasets\footnote{\url{https://github.com/huggingface/datasets}} for automatic evaluation metrics. The BART Score comes from the author's repository\footnote{\url{https://github.com/neulab/BARTScore}} and we used the one trained on ParaBank2. The hyperparameters for the experiment (non-contrastive learning) are shown in Table~\ref{tab:hparam} (applied to all models) and the ones not listed in the table are set to be default values from the transformer library. We use RAdam~\cite{radam} as the optimizer. We perform hyperparameter search on batch size from \{16, 32\}, pretrained language model learning rate from \{2e-5, 3e-5, 4e-5\}, downstream learning rate for contrastive learning from \{1e-3, 5e-4, 1e-4, 3e-5\}, negative sample composition from type A ``1 sample from each of type 1 2, filling empty case with type 3'' and type BCD ``2 samples from the type combination (2 3/1 2/1 3), filling empty case with empty string'', and the number of epochs from \{8, 15, 25\} (\{28, 32\} for  contrastive learning). For contrastive learning experiment negative sampling, we use type A for Vehicles and Food and type B for Finance. We perform our experiments on 40 GB A100 and 32 GB V100 GPU. For contrastive learning, we use batch size of 28 for Vehicles, 32 for Food, and 32 and 28 for CRA and CG respectively in Finance. The downstream learning rates are respectively 1e-3, 1e-3, 5e-4 for CRA-2+SOCL and 3e-5, 1e-4, 1e-3 for CG+SOCL in the datasets of Vehicles, Finance, and Food. We use 0.5 for the margin in the triplet loss function and 0.3 for the $\beta$ in the training loss. The experiments can take up to 10 hours. 




\subsection{Additional Model Outputs and Analysis}
\label{sec:model-output}

We present examples in Table~\ref{tab:gen-example-1}, \ref{tab:gen-example-food}, and \ref{tab:gen-example-cv}, in addition to Section~\ref{sec:quala}. From Table~\ref{tab:gen-example-1}, we can see that while BART ends the script immediately after finding an agency, our methods are able to provide detail on the interaction with the agency. This shows that training with the acquired concepts as prompts allows the model to gain more task-specific contextual knowledge, which can be potentially shared in the token embeddings of concepts. In Table~\ref{tab:gen-example-food}, the user preference in the example contains some ambiguity (i.e., it could mean making just the dough, or the complete process of making the pasta). While BART is missing the further processing on the \textit{dough}, CG and CRA-2 are able to not only include detailed and reasonable steps that process the \textit{dough} but also some information on what to do after the dough is being made, showing the usefulness of incorporating task knowledge. The reason for the improvement could be that the prompts of our method (all including the word \textit{dough}) allow the model to be attentive to the processing of $dough$. In Table~\ref{tab:gen-example-cv}, the methods without gold concepts all believe the script has ended, which is a reasonable output. Yet they would be low on automatic generation metrics because of not matching  reference well (which contains steps not necessarily needed to complete the goal). This shows the difficulty of automatically evaluating script generation since there can be many ways to solve a task and can be different in the amount of detail.

\subsection{Discussions}

\label{sec:discussion}


While we show in the experiment that TCD can be used in the script completion task in our work to achieve better performance, it can also naturally be applied to other domains such as task-oriented dialog, event causal identification, and causal inference. Furthermore, since the two different concept acquisition methods in the paper contain different advantages, it would be an interesting task to devise a method that has both benefits of being explicit in the acquisition process and being parametrized.

One natural question to ask is why we represent the KB as concept-based instead of directly using paragraphs to record each goal. While How-to websites contain human-curated solutions, the way of annotation introduces canonicalization problems since different people might have 1. different orders of completing the same task and 2. different levels of detail in the expression; on the other hand, using concepts representation helps alleviate the problem since the core concepts (e.g., ``butter'', ``temperature'', ``time'' in ``Baking a Cake'') almost always participates regardless of how they are organized in each person's memory. Furthermore, TCD introduces an interesting function of automated self-enrichment. In hierarchically related goals, such as ``how to bathe a dog'' and ``how to bathe a cat'', we can use \textbf{WordNet Hierarchy} and \textbf{Concepts Intersection} to automatically derive new nodes. We leave the experimentation to future work. 

Another important problem to be tackled is that, while human evaluation is more reliable than automatic evaluation, it is also much more costly. As mentioned in \cite{ccb_tutorial}, automatic metrics do not align with that of humans very well. One promising direction is to create an automatic metric that does not evaluate script generation methods only based on matching with the reference (since many equivalently valid references can exist).



\end{document}